\documentclass[runningheads,a4paper]{llncs}

\usepackage{amsmath}
\usepackage{amssymb}
\usepackage{graphicx}

\usepackage[latin1]{inputenc}
\usepackage[T1]{fontenc}

\usepackage{url}
\newcommand{\keywords}[1]{\par\addvspace\baselineskip
\noindent\keywordname\enspace\ignorespaces#1}

\begin{document}

\mainmatter  

\title{Using graph transformation algorithms to generate natural language equivalents of icons expressing medical concepts}

\titlerunning{Using graph transformation for generation of icon glosses}

\author{Pascal Vaillant\and Jean-Baptiste Lamy}
\authorrunning{Pascal Vaillant and Jean-Baptiste Lamy}

\institute{Universit{\'{e}} Paris 13, Sorbonne Paris Cit{\'{e}}, LIMICS, (UMRS 1142),\\
           74 rue Marcel Cachin, 93017, Bobigny cedex, France\\
           INSERM, U1142, LIMICS, 75006, Paris, France\\
           Sorbonne Universit{\'{e}}s, UPMC Univ Paris 06, UMRS 1142, LIMICS, 75006, Paris, France\\
\url{vaillant@univ-paris13.fr}}

%
%

\toctitle{Using graph transformation algorithms to generate natural language equivalents of icons expressing medical concepts}
\tocauthor{Pascal Vaillant and Jean-Baptiste Lamy}
\maketitle

\begin{abstract}
A graphical language addresses the need to communicate medical
information in a synthetic way.  Medical concepts are expressed by
icons conveying fast visual information about patients' current state
or about the known effects of drugs.  In order to increase the visual
language's acceptance and usability, a natural language generation
interface is currently developed.  In this context, this paper
describes the use of an informatics method ---graph transformation---
to prepare data consisting of concepts in an OWL-DL ontology for use
in a natural language generation component.  The OWL concept may be
considered as a star-shaped graph with a central node.  The method
transforms it into a graph representing the deep semantic structure of
a natural language phrase.  This work may be of future use in other
contexts where ontology concepts have to be mapped to half-formalized
natural language expressions.
\end{abstract}






\section{Introduction}
\label{introduction}

This work takes place in the field of medical knowledge visualization.
It is part of an ongoing project aiming at developing and promoting
the use of new interfaces for accessing medical information systems,
including a graphical representation language, VCM \cite{LamyEtAl2008},
whereby medical concepts are expressed by icons, and a multi-lingual
interface.  The graphical language is used to provide complex
information in a form adapted to synthetic visual perception.

To be accepted and used more widely within different medical
information systems, icons needs to be made as easy to learn and to
use as possible.  In this view, it is necessary to provide users with
easily accessible natural language expressions of the meaning conveyed
by any icon, e.g. in the form of a pop-up balloon appearing when the
mouse cursor is hovering over the icon.

The icons express meanings which result from the combination of a
finite set of elementary meaning components.  As there are hundreds of
thousands of potential combinations, and since the language design is
still expanding, icons are dynamically generated from the elementary
components.  Thus, the natural language utterances also have to be
automatically generated.  To this purpose, we develop a natural
language generation module, which outputs phrases in two languages.

The graphical language VCM is built against an ontology of medical
concepts, defined with the OWL-DL representation formalism
\cite{OwlGuide2004}.  Every icon expresses a concept in the VCM
ontology.  So, the primary input data is an OWL concept, which
corresponds to the medical concept to be expressed.  This one concept
is used in one process to generate an image object (not discussed
here), and in another process to generate a natural language phrase.
In the application context, more specifically, it is generally wished
that the phrase should be a noun phrase (NP); but this should be a
mere parameter of the generation process, and should it be desirable
that the output be e.g. a sentence (S), it would be possible as well.

So, the stake of the work described here is the automatic generation
of natural language expressions of concepts defined within a formal
ontology.  This problem has already raised interest in the semantic
web community \cite{Wilcock2003}, and has given way to approaches
allowing to precisely verbalize the set of logical restrictions and
specifications which define concepts in a logical description language
like OWL-DL \cite{HewlettEtAl2005}.  In the medical field, work has
been done towards automatic generation of case descriptions in natural
language from an RDF input \cite{Bontcheva2004}.

The present work adopts another approach.  It applies a method based
on the principle of graph transformation (specifically, graph grammar)
to the problem of preparing data into a form suitable for natural
language generation.  There are reasons to think that this approach is
suited to the nature of the problem, and, moreover, that it has a
potential to generalization.

\section{Background: context of use, input data}
\label{background}

\begin{figure}
\centering
\includegraphics[height=2.2cm]{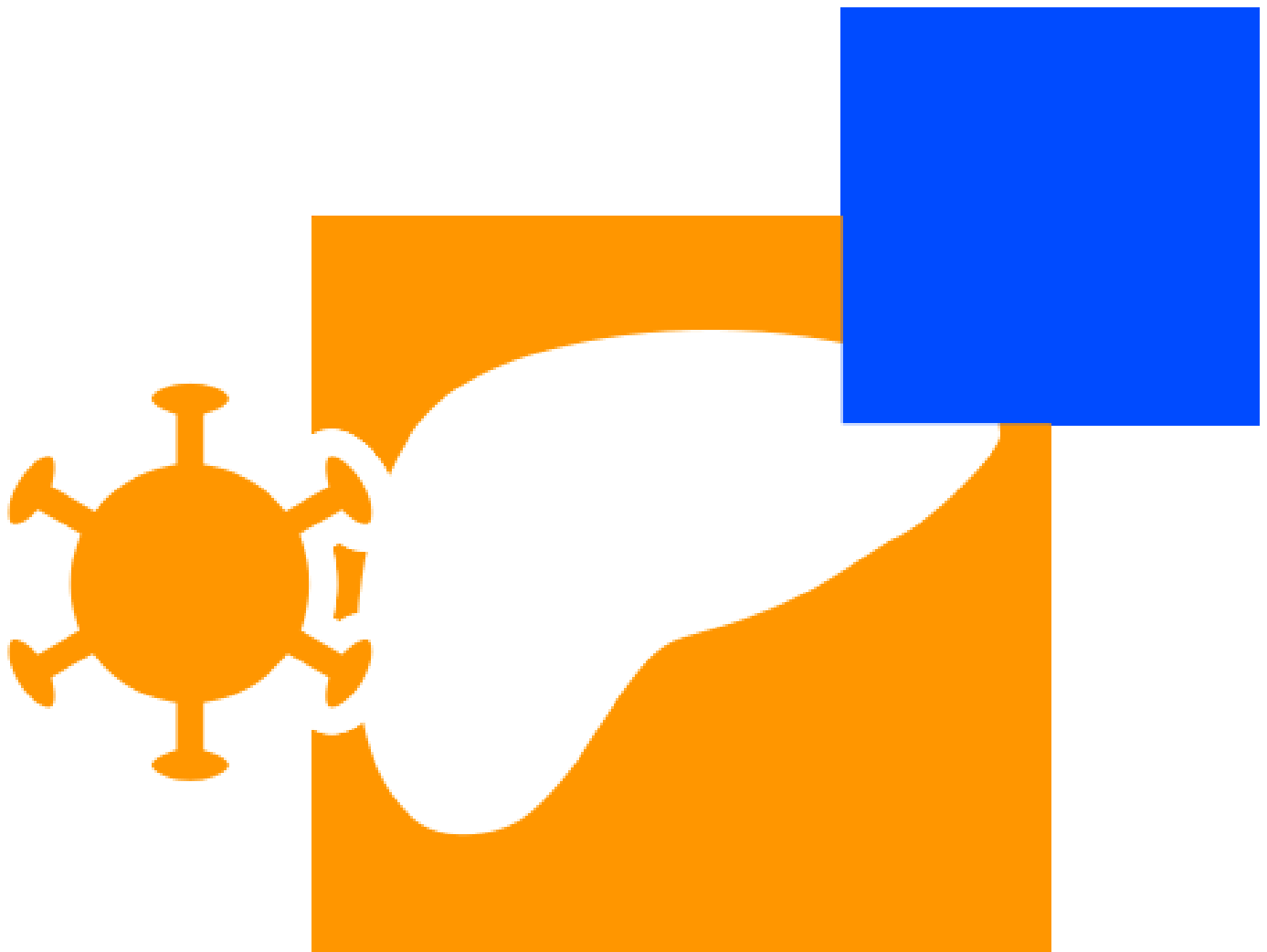}
\makebox[1cm]{~}
\includegraphics[height=2.2cm]{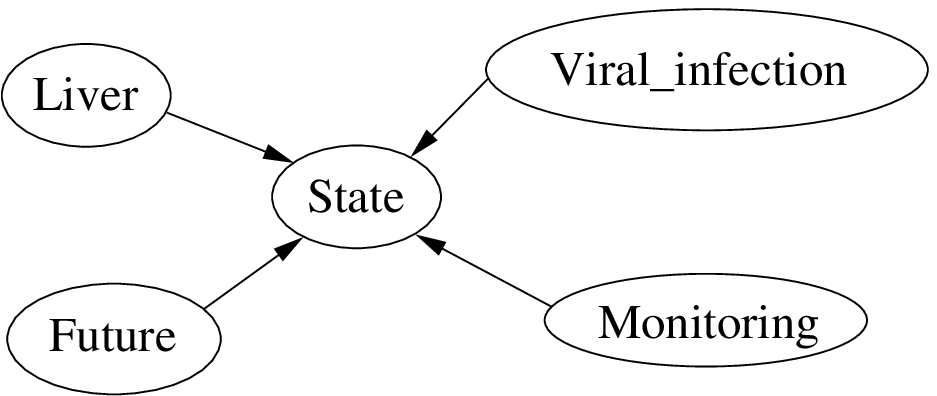}
\caption{a. Icon with internal code: {\sl risk--virus--liver--monitoring--null--null--null} (Viral hepatitis risk monitoring).
b. Graph giving the logical definition of the concept.}
\label{risk-virus-liver-monitoring}
\end{figure}

The minimal ``visual utterances'' of the graphical language are icons.
Those icons actually have a well-defined internal composition, built
against a standardized visual grammar.  The elements which make up an
icon are graphical primitives, each of which contributes to the
overall meaning of the visual sign.  An icon may for instance (Figure
\ref{risk-virus-liver-monitoring}.a) display: a central pictogram
representing a liver; embedded in a square colored in orange
(conventionally meaning ``risk''); with a shape modifier on the left
side made of a small graphical symbol representing a virus; and
another shape modifier, located in the top right corner, showing a
blue square (conventionally meaning ``monitoring'').  As an example,
such a combination of graphical primitives (central pictogram, shape,
color, side modifier, superscript modifier), with their respective
meanings (iconic or conventional) and their assembly rules, constitute
a sign conveying the concept ``viral hepatitis risk monitoring''.

Hence, the primary starting point for the text generation process is
the same as the one for the icon generation process: a standardized
code, made up of 7 positional fields, each of which corresponds to a
possible graphical primitive (cf. caption of Figure
\ref{risk-virus-liver-monitoring}).

The first step of the process actually is a parsing step.  It consists
in projecting the 7-fields code, that essentially is the specification
of a syntagm in a graphical grammar, onto an ontology of concepts.  To
this end, a dictionary defines a mapping between graphical elements
and concepts.  More precisely, every graphical element considered
within the context of a type of graphical relation between part and
whole (e.g. ``blue square {\em as superscript modifier}''), maps to
one (or more) ontology entries that define a specific property
constraining the most generic concept of the ontology, namely that of
``medical state''.  For instance, one of the rules specifies that when
a visual sign {\em has as central color} the {\em color: orange}, then
the state thus represented is linked to the {\em temporality: future}.
Similarly, if the visual sign {\em has as side modifier} the {\em
  element: virus}, then the state is linked to a {\em viral
  infection}.  Again, if the sign {\em has as central pictogram} the
{\em pictogram: liver}, then the state is linked to the {\em organ:
  liver} or to the {\em function: hepatic}.

The result of the parsing is a concept of a medical state, specified
by a number of restrictive properties, which may be represented in the
form of a graph.  For example, the icon on Figure
\ref{risk-virus-liver-monitoring}.a is a visual expression of the
graph in Figure \ref{risk-virus-liver-monitoring}.b.  Possible
ambiguities of some graphical elements (e.g. the ``liver'' pictogram
being used at times to represent the {\em organ: liver}, and at other
times to represent the {\em function: hepatic}), are removed at that
stage by a reasoner that filters valid OWL concepts.  For instance, in
the case exhibited above, it filters out the function, since a virus
may infect an organ, but not (directly) a function.

The format of the data in Figure \ref{risk-virus-liver-monitoring}.b
is not fit to be fed in to an automatic text generation process.
Natural language generation takes as input data something that should
be close to a deep semantic representation of some natural language
fragment, that is a semantic graph (we will avoid the use of the term
``conceptual graph'' coined by John Sowa \cite{Sowa1984}, since it has
a more specific formal definition; moreover, it may mislead the reader
into confusing ``concepts'' of conceptual graphs with ``concepts'' of
OWL-DL).  As a matter of fact, the graph which represents the concept
in the ontology is: (1)~of a regular shape (star-shaped);
(2)~non-ambiguous (within the reference ontology).  

On the contrary, a semantic graph should represent the semantic
structure underlying a given linguistic phrase.  Hence, it possesses
the properties expected from that level of representation, namely, it
is (1)~of an irregular shape (not necessarily star-shaped, or linear);
(2)~made up of ambiguous, i.e. multivocal, units.  The node labels in
the semantic graphs are multivocal as much in their relation to the
reference ontology (a ``semanteme'' may match more than one concept,
and a concept more than one semantemes) as in their relation to the
surface linguistic forms (a semanteme may be expressed by different
lexical units depending on its syntactic context, e.g. ``eye'' or
``ocular'').

\begin{figure}
\centering
\includegraphics[height=2cm]{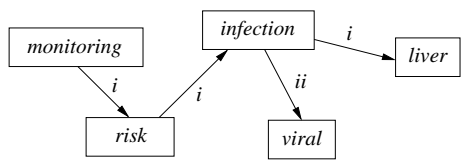}
\caption{Semantic graph corresponding to the English noun phrase ``monitoring of the risk of viral infection of the liver''.  The nodes of the graph are semantemes, the edges are weakly typed semantic relations, {\em à la} Tesnière (the lower case roman number reflects the order of ``centrality'' of the actant relative to its predicate).}
\label{semantic_graph_risk-virus-liver-monitoring}
\end{figure}

A semantic graph that would correspond to one of the most basic,
``naive'', among the many possible ways of expressing the concept of
Figure \ref{risk-virus-liver-monitoring}, would be the graph in Figure
\ref{semantic_graph_risk-virus-liver-monitoring}.

Of course, in this particular case, there are more elegant ways to
express the concept, which at the same time are closer to the actual
use by health professionals (here, ``viral hepatitis risk
monitoring'').  But we deliberately take the naive phrase as example,
because the goal of the natural language generation process described
here precisely {\em is not} to provide the most frequent term: it is
to provide a verbalization of what concept exactly is covered by the
icon, with a view to help users of health information systems better
understand the logic of the graphic language.  The idea explored in
this work is that from the same exact input code, two different
functions will generate an image on one side, and a natural language
string (or a sorted set of alternative natural language strings) on
another.

In a medical classification like e.g. the one used by ICD-10, the base
concept will be referred to by the term ``viral hepatitis'', not as
``viral infection of the liver''.  But our graphical language is
designed with the ambition to be able to express medical concepts by
combining primary elements, not to reflect an exact mapping with a
specific medical classification.  Consequences of this are: that it is
possible to build icons for concepts which are not relevant nor
frequent; that some icons might be more specific, in what they
actually express, than a most common medical term; and that other
icons might be more generic (for instance, a ``myocardial infarction''
is represented by an icon, the exact meaning of which is ``blocked
blood vessel in the heart'').  We view this independance from medical
terminology as a feature of the language.  There are many reasons for
this: first, there are different medical terminologies, and our visual
language must be able to be used in conjunction with any of them;
second, the ontology approach, with a few discrete atomic axes,
permits to express a much greater number of possible combinations of
medical concepts than the terminology approach, which sets in advance
a finite set of possibilities (this rationale also is behind the GRAIL
language, defined within the GALEN project \cite{Galen1997}).  Third,
there is no 100\% agreement between different experts on exact
mappings of even widespread medical concepts ---one more evidence that
there may be no bijective mapping between any given classification and
even a subset of a graphical language.

The initial step of the text generation process hence consists in
transforming the primary input data (the OWL concept, expressed as a
graph, like in Figure \ref{risk-virus-liver-monitoring}.b) into a deep
semantic structure (the semantic graph, like in Figure
\ref{semantic_graph_risk-virus-liver-monitoring}).  So, it is a graph
transformation process.

\section{Method: graph transformation}
\label{graph-transformation}

The problem with the preparation of data lies in the fact that some
specific configurations of properties of the initial medical concept
are jointly expressed by set words or phrases from the human language
(English in the example given here).  For example, the fact that the
medical state affects the liver (top left part of the graph in Figure
\ref{risk-virus-liver-monitoring}.b), and that it is connected with a
viral pathology (top right part), is expressed in English (among other
possibilities) by the phrase ``viral infection of the liver'' (the
present paper concentrates on the graph transformation process, and
hence does not address the issue of generating multiple possible
expressions, by using different roots ---like `hepat-' instead of
`liver'--- or by using different linguistic mechanisms ---like
morphological derivation instead of syntax).

Another property of the initial concept, e.g. in this case the fact
that it refers to a future possibility, is spontaneously expressed as
a noun phrase headed by `risk', and taking as a syntactic argument the
already built phrase (``risk of viral infection of the liver''), as in
Figure \ref{semantic_graph_risk-virus-liver-monitoring}.  This
underlying graph structure is different from the one directly drawn
from concept properties (Figure \ref{risk-virus-liver-monitoring}.b),
which, if linearized as natural language, would rather yield some text
like ``There is a risk of a state.  The state affects the liver.  The
state is related to a viral infection.''

Transformations of that type are systematic.  Changing parameters in
the entry graph would yield structurally identical phrases like
``parasitic infection of the liver'' or ``viral infection of the
respiratory tract''.

\begin{figure}
\centering
\includegraphics[height=4.21cm]{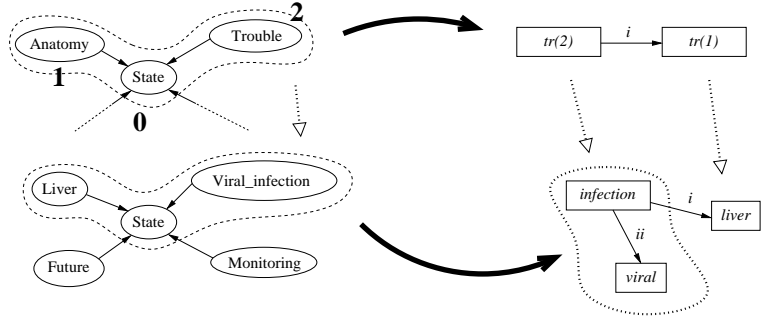}
\caption{Rewriting rule to express the generic pattern for ``trouble of an organ''.  The arab numerals as function arguments on the right side of the figure refer to boldface indices on the left side.}
\label{transf_risk-virus-liver-monitoring_1}
\end{figure}

To this end, we need graph rewriting rules allowing to specify the
systematic transformation of a sub-structure of the input graph
(corresponding to the pattern: medical state affecting an organ Y;
medical state connected to a pathology X) into a sub-structure of the
output semantic graph (corresponding to the under-specified phrase:
``{\sl <pathology Y>} of {\sl <organ X>}'').  Such a transformation
does not imply preserving either the number or the ``perimeter'' of
the nodes in the input graph when transferring their meaning into the
output graph.  For instance, the node ``Viral\_infection'' (Figure
\ref{risk-virus-liver-monitoring}.b), a unique individual entity in
the ontology, should be translated by two different semantemes of the
English language: ``viral'' and ``infection'' (Figure
\ref{transf_risk-virus-liver-monitoring_1}).

\begin{figure}
\centering
\includegraphics[height=4cm]{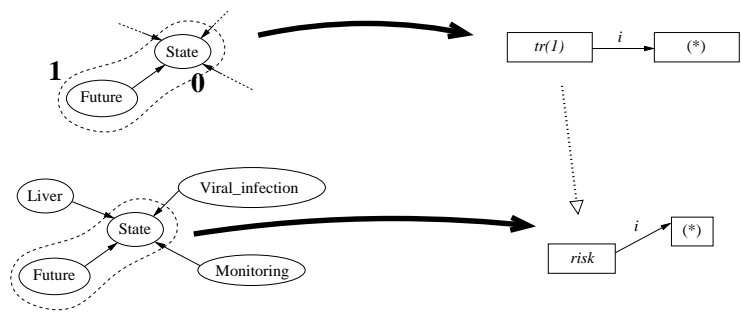}
\caption{Graph rewriting rule to express the generic pattern for ``risk of a state''.}
\label{transf_risk-virus-liver-monitoring_2}
\end{figure}

Similarly, we need a rule able to express the systematic
transformation of the subgraph in the input graph expressing the
property ``Future'' into a a subgraph in the output graph
corresponding to the lexical unit ``risk of ...'' (Figure
\ref{transf_risk-virus-liver-monitoring_2}).

There is a difference between the examples in Figure
\ref{transf_risk-virus-liver-monitoring_1} and Figure
\ref{transf_risk-virus-liver-monitoring_2}: in the first case, the
output semantic subgraph is complete, or saturated (it could be an
output graph in itself); whereas in the second case, the output
subgraph is awaiting completion by being grafted to another,
saturated, semantic graph.  The node marked with a star in the right
side of Figure \ref{transf_risk-virus-liver-monitoring_2} may be
called {\em substitution node} (in analogy with the technical term
used in the frame of the Tree-Adjoining Grammars to refer to a
comparable operation on phrase-structure trees): it is a
non-instanciated node that has to be {\em substituted} for by another
graph, given as a function argument.

The approach adopted here is based on the principle of ``graph
grammars'' \cite{EhrigEtAl1992}, which has given way to important
developments in the past two decades, especially in the field of
process modelling \cite{SchuerrEtAl1995}.  Our algorithm defines a set
of transformation rules ---or graph-rewriting rule.  Each rule takes
an under-specified graph as input, on its left-hand side ({\em filter}
subgraph), and yields another under-specified graph as output, on its
right-hand side ({\em product} subgraph).

Our algorithm may be classified in the category of {\em graph
  grammars} proper, not simply {\em graph-rewriting systems}
(following the distinction drawn by Blostein \cite{Blostein1996}),
because it makes a difference between terminal graphs and non-terminal
graphs, analogous to the similar difference that phrase-structure
grammars (PSG) make between terminal strings (made up of terminal
symbols only) and non-terminal strings in a linear language.

In the present case, a terminal graph is a graph that contains only
nodes of the type semanteme, and has no more node of the type concept.
Semantemes and concepts belong to two different XML/RDF namespaces.

\section{Method: implementation}
\label{implementation}

The generic rewriting system is implemented as a module in the {\em
  python} programming language.  It relies on four specific
mechanisms: (1)~an operation of unification of graph topological
structures, along with unification of node and edge labels; (2)~a
translation function, mapping the set of input node labels onto the
set of output node labels; (3)~a co-indexing mechanism to manage
glueing the incident edges (left loose after removing a node of the
input graph) to a node in the rewritten graph; (4)~a substitution
mechanism, defined at unsaturated nodes, to manage glueing the
neighboring (saturated) nodes to edges pertaining to the rewritten
graph.

\begin{figure}
\centering
\includegraphics[height=4cm]{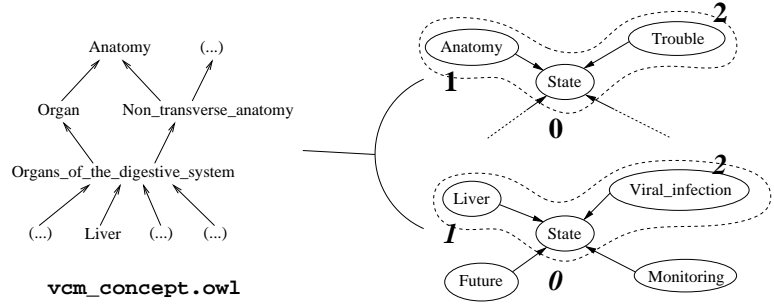}
\caption{Detecting subtype-to-supertype relations when matching a filter graph.}
\label{detecting_is_a_relations}
\end{figure}

(1)~The detection of matching sites for a filter graph (left-hand side
graph of a rewriting rule) implies: (a)~detecting an {\em isomorphism}
between part of the complete input graph and the filter graph, and
(b)~identifying {\em subtype-to-supertype} (``is a'') relations
between (more specific) node labels in the input graph and (more
generic) node labels in the filter graph.  Such ``is a'' relations
depend on the concept type hierarchy defined within the graphical
language ontology.  They allow e.g. to recognize that the subgraph
circled by a dotted line, in the bottom right corner of Figure
\ref{detecting_is_a_relations}, is a specific instance of the generic
filter graph displayer in the top right part of the same figure (by
making sure that a ``viral infection'' is a sort of ``infection'',
which is a sort of ``trouble''; and that the ``liver'' is a sort of
``organ of the digestive system'', which is a sort of ``anatomy'').

(2)~The generation of the product subgraph, when a rule is being
applied to a matching site, relies on a {\em translation} function
(noted {\sl tr} in Figures \ref{transf_risk-virus-liver-monitoring_1}
and \ref{transf_risk-virus-liver-monitoring_2}), which maps every
element in the OWL concept ontology onto a small semantic graph
(generally, but not necessarily, made up of one single semanteme
node).  In fact, since the concepts of the filter graph are
under-specified, it is not possible to specify in advance, for every
rule, the exact type of the nodes in the product subgraph.

(3)~When the filter subgraph of a rule finds a matching site on a
bigger input graph, the result of the rewriting operation is a new
graph where the subgraph found at the matching site is replaced by the
product subgraph of the rule.  The ``glueing'' of that product
subgraph with the remaining parts of the input graph relies on a {\em
  co-indexing} mechanism between product graph and filter graph.
Co-indices are attributes present on both filter-side nodes and
product-side nodes, that get numeric values; when a filter-side node
and a product-side node share the same co-index, it means that they
should match the same node in the input graph.  The actual integer
number used as value for a co-index in the definition of a rule may be
arbitrary: its only purpose is to be shared by the left-hand side and
the right-hand side.  If there are more than one co-index, different
integer values mean that the relevant nodes should match distinct
nodes in the input graph.  Hence, co-indices allow to spot the nodes
in the input graph where loose incident relations of the product
subgraph have to be ``glued''.

\begin{figure}
\centering
\includegraphics[height=1.8cm]{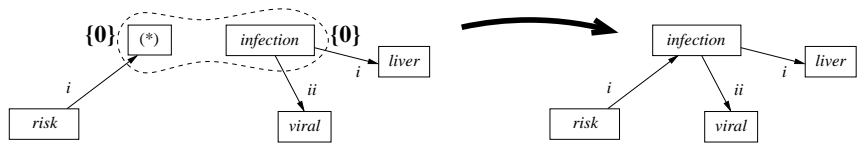}
\caption{Substitution of an unspecified node in one graph by a saturated node in another graph.}
\label{subst_risk-virus-liver-monitoring}
\end{figure}

(4)~Some product subgraphs are made up of a set of fully determined
semanteme-nodes, that express all the concepts of the input graph
which were captured when matching the filter subgraph (Figure
\ref{transf_risk-virus-liver-monitoring_1}).  Other, oppositely, have
a loose edge ---to put it another way, they include an edge between a
node which is already fully determined in the product subgraph, and a
node which has to be determined somewhere else (Figure
\ref{transf_risk-virus-liver-monitoring_2}).  Such product subgraphs
contain a {\em substitution} node.  After the application of the rule,
the substitution node must be unified with a saturated node from the
remaining of the graph, to build the whole rewritten graph (Figure
\ref{subst_risk-virus-liver-monitoring}).  Substitution is compulsory.

Remark: Points (3) and (4) actually are implemented by the same
underlying computer function operating on graphs, and taking two
arguments: the ``graft'' and the ``trunk''.  This function attempts to
find co-indexed nodes on both sides sharing the same value, and it
``glues'' the two graphs on those nodes.  For every such ``co-indexed
site'', one of the sides must be filled and the other side blank.
(3)~is implemented when the trunk node is blank and the graft node is
filled; (4)~is implemented when it is the other way around.

The algorithm works by iterated rewritings:

At step $0$, the set of rewritings ${\boldsymbol{R}}_{0}$ is
initialized to G; G being the original input graph, representing the
OWL concept.

At every step $n+1$, ${\boldsymbol{R}}_{n+1}$ is augmented with the
set of rewritings yielded by applying matching rules to elements of
${\boldsymbol{R}}_{n}$, when those results do not already belong to
${\boldsymbol{R}}_{n}$:

\begin{displaymath}
{\boldsymbol{R}}_{n+1} \leftarrow{} {\boldsymbol{R}}_{n} \cup{} \{ {{\boldsymbol{R}}_{k}}(g)\ |\ g \in{} {\boldsymbol{R}}_{n} \} ,
\end{displaymath}

where ${{\boldsymbol{R}}_{k}}(g)$ denotes the result of rewriting a
graph $g$ (present in ${\boldsymbol{R}}_{n}$) by one of the applicable
rewriting rules, $k$.

When the set ${\boldsymbol{R}}_{n}$ ceases to grow between two
iterations, the loop is exited, and ${\boldsymbol{R}}_{n}$ is filtered
so that only the ``terminal'' graphs are kept (the graphs where all
the nodes are semantemes, and no more concepts).

In our system, the generic processing mechanisms are separated from
the description of specific rewriting rules, like it is common
practice in the field of formal grammars (it is an instance of the
more general principle that data should be treated separately from
processes).  The former are implemented by functions in the {\em
  python} programming language, taking graph-rewriting rule
identifiers as en input parameter; the latter are stored in XML
documents following an {\em ad hoc} document schema.

\section{Conclusion and Perspective}
\label{conclusion}

The next step in the present work is the development of a complete
text generation module, based on the generation of phrase structure
trees by derivation of elementary trees in a TAG lexicalized grammar
\cite{SchabesAbeilleJoshi1988}.

The graphical language is built on minimal segments of expression
called {\em icons}, a description of which has been given above
(Section \ref{background}).  Those icons may be combined together,
following a constrained visual syntax, to compose more complex iconic
utterances: on bidimensional surfaces, structured in predefined
fields, they form synthetic visualization grids displaying information
about the complete set of contraindications or side effects of a drug,
or the clinical condition of a patient.

A future extension of the natural language generation work will be
taking into account that visual syntax, to be able to translate
complex graphical utterances in texts in the chosen target natural
language.  It is envisioned that future developments shall include
other output languages, so that the visual language approach actually
allows embedding in multi-lingual systems for displaying medical
information.

We believe that the method presented here has a potential for
generalization.  It can be used in other cases where generation of
natural language equivalents of OWL concepts may be desirable as a
tool to help ontology users; and, more generally, when the
pre-linguistic input for natural language generation is expressed in a
knowledge representation formalism translatable in the form of graphs.
This might be of use in other application fields, like automatic
explanation generation in health information systems, or help in
decision making.

\end{document}